\documentclass{article}

\usepackage{arxiv}

\usepackage[utf8]{inputenc} 
\usepackage[T1]{fontenc}    
\usepackage{hyperref}       
\usepackage{url}            
\usepackage{booktabs}       
\usepackage{amsfonts}       
\usepackage{nicefrac}       
\usepackage{microtype}      
\usepackage{lipsum}
\usepackage{graphicx}
\usepackage{natbib}
\usepackage[load-configurations=version-1]{siunitx} 

\title{ECGDetect: Detecting Ischemia via Deep Learning}

\author{
  Atandra Burman\\
  RCE\\
  \texttt{atandra@rce.ai} \\
   \And
 Jitto Titus, PhD\\
  RCE\\
  \texttt{jitto@rce.ai} \\
   \And
  David Gbadebo, MD \\
  East Atlanta Cardiology\\
  \texttt{tdgbadebo@gmail.com} \\
     \And
 Melissa Burman, MD \\
  Northside Hospital Gwinnett, GA \\
  \texttt{melissa.burman1@gmail.com} \\
}

\begin{document}
\maketitle

\begin{abstract}
Coronary artery disease(CAD) is the most common type of heart disease and the leading cause of death worldwide[1]. A progressive state of this disease marked by plaque rupture and clot formation in the coronary arteries, also known as an acute coronary syndrome (ACS), is a condition of the heart associated with sudden, reduced blood flow caused due to partial or full occlusion of coronary vasculature that normally perfuses the myocardium and nerve bundles, compromising the proper functioning of the heart. Often manifesting with pain or tightness in the chest as the second most common cause of emergency department visits in the United States, it is imperative to detect ACS at the earliest.
A classic hallmark of early stages in ACS is myocardial ischemia recognized by patient symptoms, coupled with morphological patterns in ECG. In specific, ST elevation on the ECG is presented in up to 25\% of ACS patients and the rest (non-ST elevation-ACS (NSTE-ACS) or unstable angina (UA)) show non-specific ECG changes. The 75\% of ACS patients at risk for transient myocardial ischemia, can be detected with ambulatory ECG monitoring, allowing the early detection and clinical intervention towards a better patient prognosis. Identifying myocardial ischemia early helps reduce irreversible damage to cardiac tissues and prevent patient deterioration. This is particularly relevant to diabetic patients at home, that may not feel classic chest pain symptoms, and are susceptible to silent myocardial injury. In this study, we developed the RCE- ECG-Detect algorithm, a machine learning model to detect the morphological patterns in significant ST change associated with myocardial ischemia. We developed the RCE- ECG-Detect using data from the LTST database which has a sufficiently large sample set to train a reliable model. We validated the predictive performance of the machine learning model on a holdout test set collected using RCE's ECG wearable. Our deep neural network model, equipped with convolution layers, achieves 90.31\% ROC-AUC, 89.34\% sensitivity, 87.81\% specificity. 

\end{abstract}

\section{Introduction}
Problem: Heart attack or Myocardial Infarction (MI) is one of the leading causes of death around the world. Myocardial ischemia is an early phase in coronary insufficiency that can be reversed if detected early, and in time for revascularization, thereby enabling prevention of MIs. 
Outcome of the potential solution: If 1\% of the ischemia can be detected, that would lead to thousands of lives being saved. 
Solution: RCE has developed a ECG wearable in the form of machine washable inner vest with embedded dry electrodes that continuously collect 12 channel ECG waves. This continuous monitoring can be used to train machine learning algorithms to detect ischemia. 
Many different types of cardiac diseases including myocardial infarction, ventricular tachycardia, and atrial fibrillation can be diagnosed from ECG signals as ECG signals are easily available. Ischemia detection from ECG recordings is usually performed by expert technicians and cardiologists given the complex nature of the ECG signal. 
In order to detect ischemia, an algorithm must implicitly recognize all different types of waves with an ECG and decouple complex relationships between them over time. ECG signals battery from patient to patient, and often presents with diverse noise artifacts, thus increasing the complexity of interpretation and detection of ECG morphology.

\section{Method} 

In the following section, we first describe the datasets used in this study for training and testing purposes. Then we provide a detailed description of the neural network model which is used for the classification of normal vs ischemic ECG. 
\subsection{Data Set}

\subsubsection{LTST Database}

We used a Long-term ST database from Physionet to create a training dataset. The LTST database contains 20-24 hour ambulatory 2-3 lead ECG recordings. These recordings are sampled at 250Hz. We re-sample these ECG data to 200Hz for our analysis. The annotation provided in the LTST database is semi-automatically generated by human experts and computer algorithms. More information about the annotations can be accessed here[2].  It is worth noting that there are three different levels of ST change annotation available in the LTST database. We use protocol-B where V-Min is set to 100uv and T-min is set to 30seconds. 

\subsubsection{RCE-India Data collection protocol}

\textbf{Study design}
The objective of the research was to perform automated ischemia detection using ECG waveforms analyzed by deep learning.
\begin{itemize}
    \item  Number of patients collected: 14 ischemic, 14 non-ischemic
\item Age group: 55-85
\item Inclusion: 
 Cohort 1 - Patients evaluated for chest pain in cardiac observation unit
 Cohort 2 - Non-cardiac risk patients admitted to general ward
\item De-identified ECG samples were collected from patients at a community hospital and cardiac research institute in Nagpur, India. 
\item No identifiable patient information was collected and shared with the research team.
\item IRB: yes; Ethics approval and Consent to Participate.
\end{itemize}

\subsection{RCE ECG Vest}
Currently, there is a lack of a remote monitoring solution effectively focussed towards acquiring and analyzing ECG during early ischemia. RCE has solved this with a wearable that positions 10 electrodes in the back of the body, conforming to a stable back musculature ensuring reliable and effective contact with the body. The relative locations of the 10 electrodes mirror the ECG generated from conventional frontal limb leads and pre-cordial leads. 

The dry electrodes and leads are integrated with comfortable wearable fabric that does not require application of wet gel or cleaning, and is robust enough to sustain machine washes. This wearable designed as an inner vest allows seamless adoption in the lifestyle of elderly demographic while driving value towards consistent and reliable data acquisition.

\section{Problem Formulation}
In this section, we describe our formulation of this task. The ECG ischemia detection task is a classification task which takes as input an ECG signal X = [x1,..xk] and outputs a sequence of labels y.  such that each $y_i$ can provide the probability of ischemia. Each output label corresponds to a segment of the ECG x(i). We optimize the cross-entropy objective function - a neural network used for the classification of different types of ischemia from the ECG dataset. The neural network which is used in this study is a convolutional neural network. We do not describe the basics of a convolutional neural network and have provided references for readers to further read.

\begin{figure*}
  \includegraphics[width=1\textwidth]{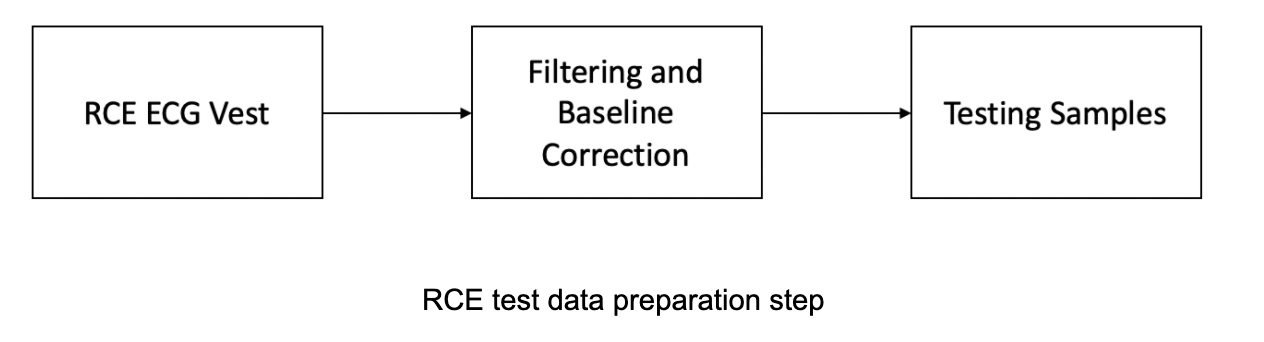}
  \centering
\end{figure*}

\section{Model Architecture and Training}
We frame the analysis in our paper as a classification task, where we use ECG as the input for the model and classify the ECG as normal vs ischemic. This binary classification setting considers different labels such as STEMI, NSTEMI as ischemic ECG, and rest as normal ECG. In order to make the optimization of such a network tractable, we employ shortcut connections in a similar manner to those found in the Residual Network architecture\cite{he2016deep}. The shortcut connections between neural network layers optimize training by allowing information to propagate well in very deep neural networks. Before the input is fed into the network, it is normalized using a robust normalization strategy. The network consists of 16 residual blocks with 2 convolutional layers per block. The convolutional layers all have a filter length of 16 and have 64k filters, where k starts out as 1 and is incremented every 4-th residual block. Every alternate residual block subsamples its inputs by a factor of 2, thus the original input is ultimately subsampled by a factor of 2 8 . When a residual block subsamples the input, the corresponding shortcut connections also subsample their input using a Max Pooling operation with the same subsample factor.
Before each convolutional layer we apply Batch Normalization  and a rectified linear activation, adopting the pre-activation block design. The first and last layers of the network are specially caused due to this pre-activation block structure. We also apply Dropout between the convolutional layers and after the non-linearity. The final fully connected layer and softmax activation produce a distribution over the 14 output classes for each time-step\cite{schirrmeister2017deep}.
We train the networks from scratch, initializing the weights of the convolutional layers as in. We use the Adam  optimizer with the default parameters and reduce the learning rate by a factor of 10 when the validation loss stops improving. We save the best model as evaluated on the validation set during the optimization process.

\begin{figure*}
  \includegraphics[width=1\textwidth]{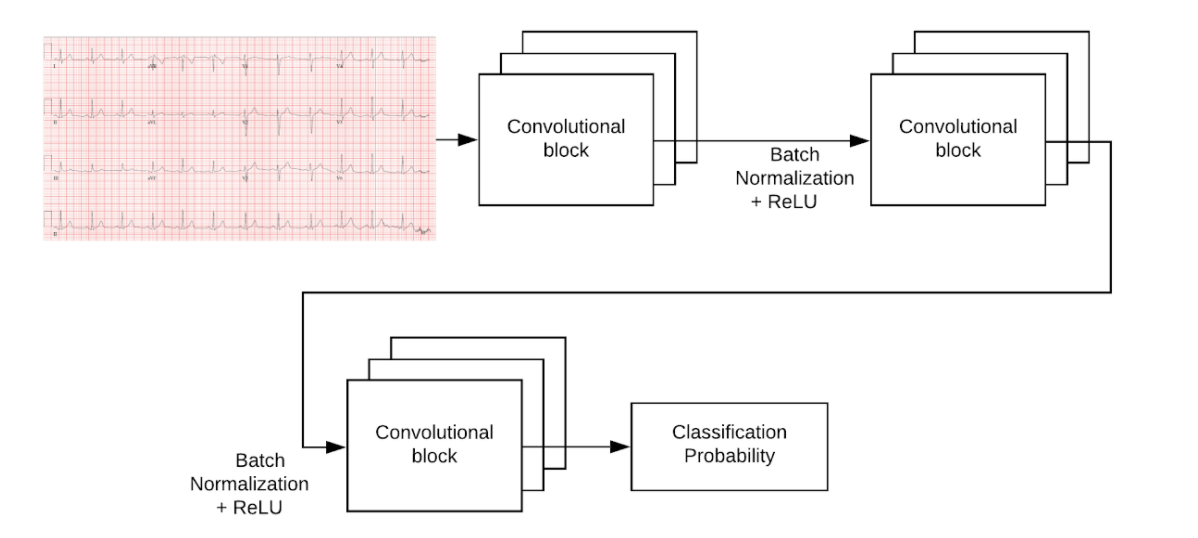}
  \centering
\end{figure*}

\subsection{Testing} 
We collected a test set of 573 records from 28 unique patients. The ground truth annotations for each patient were carefully selected after reviewing the data collected. We apply baseline correction and filtering to preprocess the ECG waveforms. 

\section{Experiments and Results}
In this study, we developed the RCE- ECG-Detect algorithm, a machine learning model to detect the significant ST Change. We developed the RCE- ECG-Detect using data from the LTST database which has a sufficiently large sample set to train a reliable model. We validated the predictive performance of the machine learning model on a holdout test set collected using RCE vest.

\subsection{Baselines: Comparison with Traditional Approaches}
We also compare the performance of our proposed model to that of the best performing traditional approach where typically features are extracted from the ECG waveform and used with a classification algorithm such as Logistic Regression, Random Forest, Gradient Boosting algorithm.

\subsection{Results}
In our analysis, we asked a few questions to obtain a better understanding of the model. The results and analysis to these questions are provided below:  

\begin{itemize}
    \item \textbf{Is the model trained on LTST data performs well on RCE ECG data?}: We tested the trained model on RCE ECG data to measure performance of our trained model.  The results of this experiment are presented in table 1. We show that our trained model outperforms baseline models such as logistic regression and random forest. Our trained model CNN was able to achieve ROC-AUC of 91.56\%, 89.53\%, 88.71\%  specificity.  This shows that our model is able to classify ischemia well. We have provided different hyperparameters in the supplementary which were used in tuning these algorithms to obtain these performances.

\item\textbf{How well the model performs in the held out LTST dataset?}: We also evaluate the performance of our model by validating the trained model on heldout test sets from LTST database. This experiment was designed to understand the efficacy of the trained model 

\item\textbf{Is the model able to perform at a similar level of expert cardiologists?}:
While current test sets for ischemic and non-ischemic ECG data sets were small in running a comprehensive review, anecdotal cases were verified with an unbiased board certified EP cardiologist. As we continue further data collection efforts, we intend to conduct a comparative between model performance and multiple expert cardiologists. 
\end{itemize}

\section{Conclusion}

RCE’s initial model training gives us line of sight towards the value of ischemia detection using AI model coupled with ambulatory ECG wearable that can be used by patients at homes. The application of the RCE-ECG-Detect algorithm, has particular relevance to the diabetic demographic with high risk coronary artery disease with sequelae of neuropathy, who are susceptible to silent MIs.

\newpage
\bibliography{sample}

\end{document}